\documentclass[letterpaper]{article} 
\usepackage{aaai2026}  
\usepackage{times}  
\usepackage{helvet}  
\usepackage{courier}  
\usepackage[hyphens]{url}  
\usepackage{graphicx} 
\usepackage{natbib}  
\usepackage{caption} 
\usepackage{algorithm}
\usepackage{algorithmic}
\usepackage{amssymb}
\usepackage{amsmath}
\usepackage{booktabs}
\usepackage{multirow}
\usepackage{makecell}
\usepackage{hhline}
\usepackage{pifont}
\usepackage{newfloat}
\usepackage{listings}
\usepackage{xcolor}

\usepackage{colortbl}   

\DeclareCaptionStyle{ruled}{labelfont=normalfont,labelsep=colon,strut=off} 
\floatstyle{ruled}
\newfloat{listing}{tb}{lst}{}
\floatname{listing}{Listing}
\title{Signal: Selective Interaction and Global-local Alignment for Multi-Modal Object Re-Identification}
\author{
    Yangyang Liu\textsuperscript{\rm 1}\equalcontrib,
    Yuhao Wang\textsuperscript{\rm 1}\equalcontrib,
    Pingping Zhang\textsuperscript{\rm 1,\rm 2}\thanks{Corresponding author.}
}
\affiliations{
    \textsuperscript{\rm 1}School of Future Technology, Dalian University of Technology\\
    \textsuperscript{\rm 2} National Engineering Laboratory for Integrated Aero-Space-Ground-Ocean Big Data Application Technology, Xi'an, China\\
    \{yyliu, 924973292\}@mail.dlut.edu.cn,zhpp@dlut.edu.cn,
}

\usepackage{bibentry}

\begin{document}

\maketitle

\begin{abstract}
Multi-modal object Re-IDentification (ReID) is devoted to retrieving specific objects through the exploitation of complementary multi-modal image information.
Existing methods mainly concentrate on the fusion of multi-modal features, yet neglecting the background interference.
Besides, current multi-modal fusion methods often focus on aligning modality pairs but suffer from multi-modal consistency alignment.
To address these issues, we propose a novel selective interaction and global-local alignment framework called \textbf{Signal} for multi-modal object ReID.
Specifically, we first propose a Selective Interaction Module (SIM) to select important patch tokens with intra-modal and inter-modal information.
These important patch tokens engage in the interaction with class tokens, thereby yielding more discriminative features.
Then, we propose a Global Alignment Module (GAM) to simultaneously align multi-modal features by minimizing the volume of 3D polyhedra in the gramian space.
Meanwhile, we propose a Local Alignment Module (LAM) to align local features in a shift-aware manner.
With these modules, our proposed framework could extract more discriminative features for object ReID.
Extensive experiments on three multi-modal object ReID benchmarks (i.e., RGBNT201, RGBNT100, MSVR310) validate the effectiveness of our method.
The source code is available at https://github.com/010129/Signal.
\end{abstract}
\section{Introduction}
\begin{figure}[t]
    \centering
    \includegraphics[width=1.0\columnwidth]{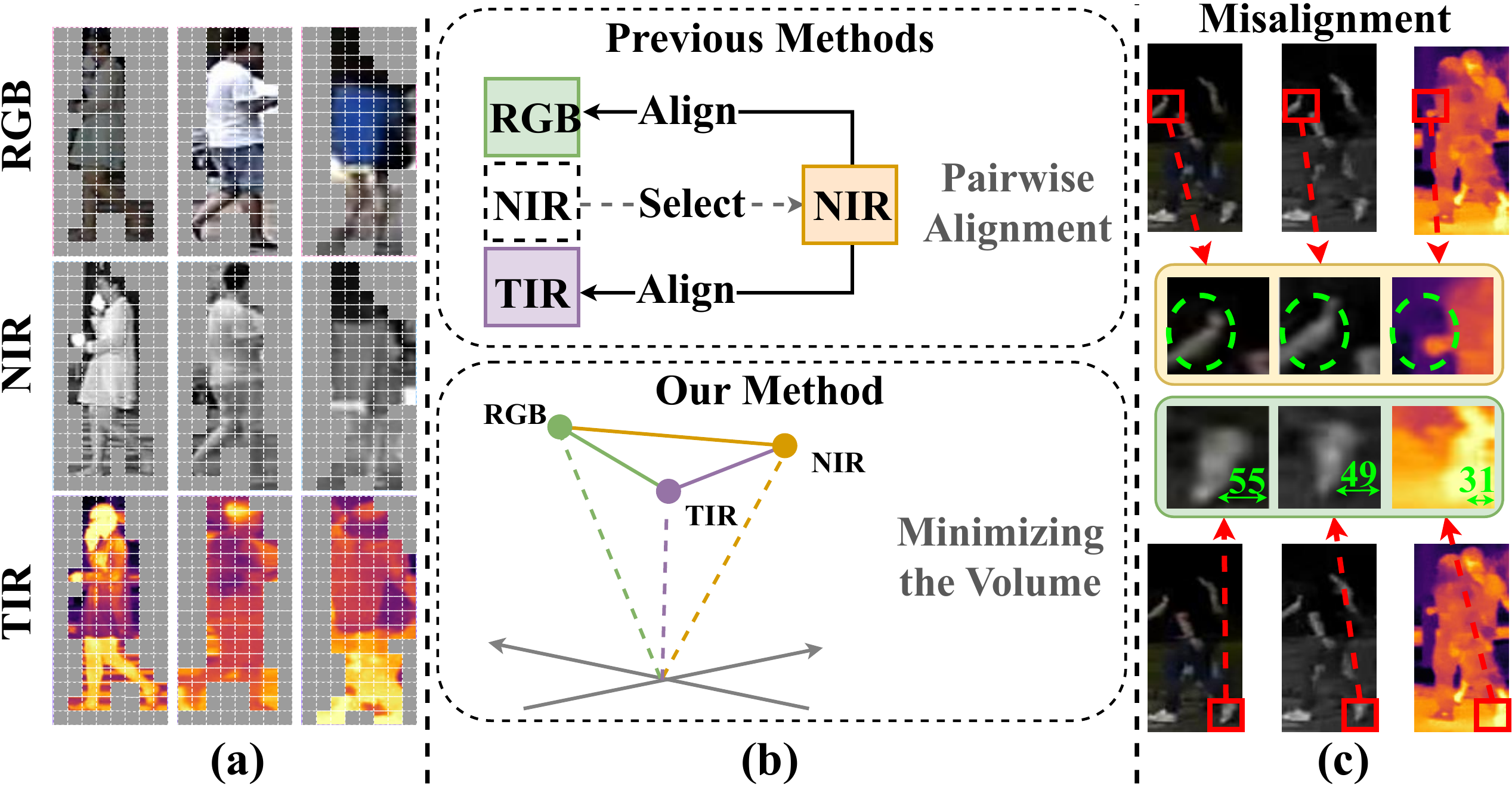}
    \caption{
    Motivations of our framework.
    (a) Background interferences in multi-modal images.
    (b) Comparison between previous pairwise alignment and our simultaneous alignment.
    (c) Misalignments exist across different modalities.}
    \label{fig:introduction}
\end{figure}

Object Re-IDentification (ReID) aims to retrieve identical objects across non-overlapping cameras.
Initially, researchers focus on single-modal object ReID~\cite{he2021transreid,zhang2021hat,liu2021watching} mainly based on RGB images.
However, adverse environments such as darkness and strong light can cause blurry details in RGB images.
Later, researchers find that Near Infrared (NIR) and Thermal Infrared (TIR) images exhibit strong robustness in harsh visual environments.
With complementary information from different modalities, existing multi-modal object ReID methods~\cite{lin2025dmpt,wan2025reliable,li2025icpl,feng2025multi,li2025next,wan2025ugg} achieve outstanding performance.
However, they ignore background interference in each modality.
As shown in Fig.~\ref{fig:introduction} (a), irrelevant background information may introduce noise and affect the extraction of discriminative information.
Meanwhile, recent multi-modal fusion methods~\cite{wang2023connecting,yu2024mm,yang2024enhancing} focus on aligning different modalities in a pairwise manner through contrastive learning.
As illustrated in the upper part of Fig.~\ref{fig:introduction} (b), these methods typically select one modality as the anchor and align the remaining modalities to it.
However, this pairwise alignment strategy becomes less effective when scaling to more than two modalities, as it fails to capture the complex relationships among all modality pairs.
Thus, as shown in the lower part of Fig.~\ref{fig:introduction} (b), simultaneously aligning multiple modalities without relying on a fixed anchor modality offers a more flexible solution for multi-modal alignment.
Besides, current multi-modal imaging sensors often struggle to ensure precise pixel-level alignment.
As shown in Fig.~\ref{fig:introduction} (c), pixel-level misalignment commonly exists across different modalities, leading to semantic inconsistency in multi-modal fusion.

Motivated by the aforementioned observations, we propose \textbf{Signal}, a novel selective interaction and global-local alignment framework for multi-modal object ReID.
Our proposed framework comprises three components: the Selective Interaction Module (SIM), the Global Alignment Module (GAM) and the Local Alignment Module (LAM).
First, SIM selects important patch tokens from multi-modal features by evaluating their significance both within and across modalities.
This is achieved by computing intra-modal and inter-modal attention scores to guide the selection process.
Second, GAM enables the simultaneous alignment of multi-modal features by minimizing the volume of 3D polyhedra in the gramian space~\cite{cicchetti2025gramian}, as shown in the lower part of Fig.~\ref{fig:introduction} (b).
Unlike existing pairwise alignment methods, our method eliminates the need for a fixed anchor modality, enabling a more flexible and effective alignment across multiple modalities.
Third, LAM further refines this process by focusing on fine-grained alignment at the local feature level.
Leveraging deformable sampling, LAM adaptively aligns local details across modalities in a shift-aware manner, mitigating semantic inconsistency caused by pixel-level misalignment.
With the above components, our proposed framework effectively addresses the challenges of background interference and multi-modal misalignment for robust multi-modal feature learning.
Extensive experiments on three multi-modal object ReID datasets validate our method's effectiveness.

Our main contributions are summarized as follows:
\begin{itemize}
    \item
    We propose a novel selective interaction and global-local alignment framework named Signal for multi-modal object ReID, which effectively addresses the challenges of background interference and multi-modal misalignment.
    \item
    We propose the Selective Interaction Module (SIM) to leverage inter-modal and intra-modal attention scores for selecting important patch tokens, thereby mitigating background interference in multi-modal fusion.
    \item
    We propose the Global Alignment Module (GAM) to  simultaneously align multi-modal features through minimizing the volume of 3D polyhedra in the gramian space.
    \item
    We propose the Local Alignment Module (LAM) to align local features in a shift-aware manner, effectively addressing pixel-level misalignment across modalities.
    \item
    Extensive experiments on three multi-modal object ReID datasets validate the effectiveness of our method.
\end{itemize}
\section{Related Work}
\subsection{Multi-Modal Object Re-Identification}
Multi-modal object ReID is devoted to retrieving specific objects through the exploitation of multi-modal inputs.
Existing methods focus on learning complementary image features.
For example, Wang \emph{et al.}~\cite{wang2024topreid} propose a cyclic token permutation framework to reduce the distribution gap across different modalities.
Feng \emph{et al.}~\cite{feng2025multi} integrate pixel-level interaction to balance modality-specific features.
Wang \emph{et al.}~\cite{wang2025demo} introduce the mixture of experts for adaptive weighting decoupled features.
Besides, researchers find that graph-based models exhibit superior capabilities in modeling complex relational structures.
Thus, Wan \emph{et al.}~\cite{wan2025reliable} introduce graph inference with modality awareness for improving feature robustness.
Wan \emph{et al.}~\cite{wan2025ugg} further quantify uncertainty through graph models.
Recently, researchers start to explore the use of Multi-modal Large Language Models (MLLMs) to enhance multi-modal feature learning.
For instance, Wang \emph{et al.}~\cite{wang2025idea} integrate semantic guidance from inverted texts generated by MLLMs.
Li \emph{et al.}~\cite{li2025next} introduce text-modulated and context-shared experts to enhance feature robustness.
However, these methods primarily focus on feature fusion and ignore the background interference in multi-modal object ReID.
To address this issue, Zhang \emph{et al.}~\cite{zhang2024magic} propose the object-centric feature refinement to mitigate background interference.
Zhang \emph{et al.}~\cite{zhang2025promptma} introduce token selection to filter out the irrelevant background noise.
Although the above methods achieve remarkable performance, they typically select tokens within each modality separately, ignoring the importance of tokens across modalities.
Meanwhile, previous methods~\cite{wang2024topreid,zhang2024magic} mainly focus on pairwise alignment, which exhibits limitations in complex multi-modal scenarios.
Thus, we introduce the selective interaction with intra-modal and inter-modal attention scores to mitigate background interference.
In addition, we perform multi-modal alignment in the gramian space, which offers a great flexibility compared with pairwise alignments.
\begin{figure*}[t]
\centering
\includegraphics[width=0.94\textwidth]{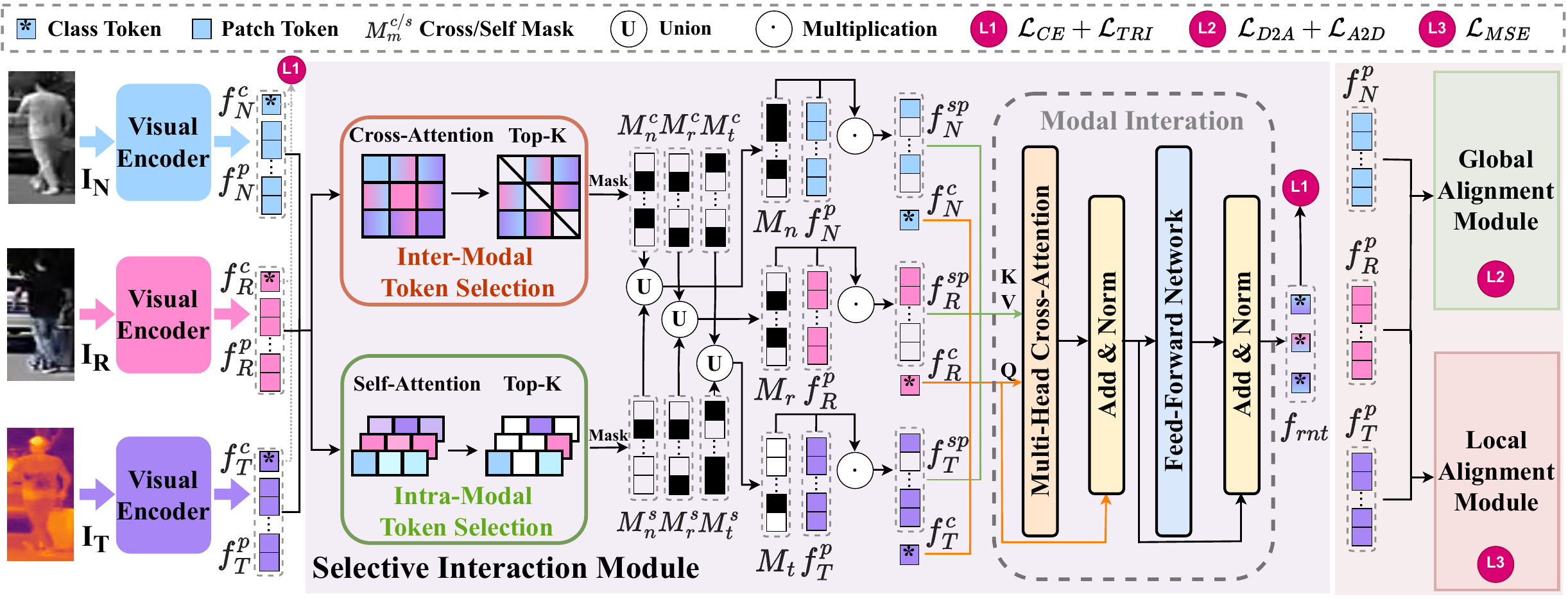}
\caption{Illustration of our proposed framework.}
\label{fig:Signal}
\end{figure*}
\subsection{Multi-Modal Feature Fusion}
Multi-modal feature fusion leverages complementary information from different modalities to enhance feature robustness.
For example, Li \emph{et al.}~\cite{li2025mulfscap} propose a learnable modality dictionary to preserve consistency between individual modality features.
Dai \emph{et al.}~\cite{dai2025novel} enhance cross-modal feature fusion through contrastive learning and reduce redundancy by utilizing visual sequence compression.
Additionally, Nagrani \emph{et al.}~\cite{nagrani2021attention} employ fusion bottlenecks to facilitate modality information aggregation.
In multi-modal object ReID, Zhang \emph{et al.}~\cite{zhang2024mambareid} enhance feature discrimination by integrating inter-modality information with shallow and deep features through dense connections.
Wang \emph{et al.}~\cite{wang2025mambapro} propose a synergistic residual prompt to guide the joint learning of multi-modal features.
Following this direction, later studies increasingly focus on effective feature alignment across modalities.
For instance, Wang \emph{et al.}~\cite{wang2024topreid} utilize complementary reconstruction to minimize the distribution gap across different modalities.
Zhang \emph{et al.}~\cite{zhang2024magic} introduce a pairwise background consistency constraint to align background features across modalities for improved feature representation.
Although these methods achieve remarkable performance, the simultaneous alignment of multi-modal features remains an under-explored area in current work.
To bridge this gap, we propose a novel multi-modal alignment in the gramian space.
It enables holistic and anchor-free alignment by modeling the global interactions among all modalities simultaneously in a unified space.
\section{Methodology}
As shown in Fig.~\ref{fig:Signal}, our proposed framework is consist of the Selective Interaction Module (SIM), the Global Alignment Module (GAM) and the Local Alignment Module (LAM).
In this section, we will describe the details of each module.
\subsection{Selective Interaction Module}
Background information often interferes with multi-modal object ReID.
Existing methods~\cite{wang2025idea,li2025icpl,wang2025mambapro,wan2025ugg} lack an effective way to eliminate the background interference.
To address this issue, we propose a Selective Interaction Module (SIM) to select important patch tokens with intra-modal and inter-modal information.
Specifically, a given image $I \in \mathbb{R}^{3 \times H \times W}$ is first split into $L$ patches, where $H$ and $W$ denote the height and width, respectively.
Then, the patch tokens are fed into a visual encoder to extract modality-specific features.
As a result, the image features $\mathcal{F}_{m} \in \mathbb{R}^{(L+1) \times D}$ for each modality can be expressed as follows:
\begin{equation}
\mathcal{F}_{m} = \{f_m^{\text{cls}}; f_m^1, f_m^2, \cdots, f_m^L\}, \quad m \in \{N, R, T\},
\end{equation}
where $f_m^{\text{cls}}$ denotes the [CLS] token and $f_m^i$ represents the token of the $i$-th patch.
Here, $N$, $R$ and $T$ correspond to the NIR, RGB and TIR modalities, respectively.
To facilitate better explanations, we denote the features separately as:
\begin{equation}
f_m^c = f_m^{cls},
\end{equation}
\begin{equation}
f_m^p =  \{f_m^1, f_m^2, \cdots, f_m^L\}.
\end{equation}
Here, $f_m^c \in \mathbb{R}^{D}$, $f_m^p \in \mathbb{R}^{L \times D}$ and $D$ is the embedding dimension.
Then, to facilitate the selection of important patches within each modality, we propose the Intra-Modal Token Selection and Inter-Modal Token Selection.

\textbf{Intra-Modal Token Selection.}
To assess the importance of patches within each modality, we introduce the Intra-Modal Token Selection.
It leverages self-attention scores to preliminarily select and retain the most informative patches.
More specifically, we first compute the attention scores of all patch tokens within each modality as follows:
\begin{equation}
Q_m = f^{c}_m W_q, K_m = f^{p}_m W_k,
\end{equation}
\begin{equation}
\mathcal{S}_m = \mathrm{Softmax}\left(\frac{Q_m K_m^\top}{\sqrt{D}}\right),
\end{equation}
where \( W_q \) and \( W_k \) are identity matrices and they yield superior performance.
Here, $\mathcal{S}_m \in \mathbb{R}^{1 \times L}$ denotes the self-attention score of each patch token in modality $m$.
Next, we select the high-similarity patches within each modality from $\mathcal{S}_m$ as follows:
\begin{equation}
\Theta_m = TopK(\mathcal{S}_m,k_1).
\end{equation}
Here, \( k_1 \) denotes the number of important patch tokens to be selected.
\( \Theta_m \) is the index set of the top-\( k_1 \) patches selected within each modality.
We then construct a binary mask using \( \Theta_m \) to retain the selected patch tokens:
\begin{equation}
M_m^s = \varPsi(\Theta_m).
\end{equation}
where \( \varPsi \) denotes a binary masking operation.
Each element in \( M_m^s \) indicates whether the patch token is selected (1) or discarded (0).
Through this process, we obtain a preliminary set of intra-modality important patches for each modality.

\textbf{Inter-Modal Token Selection.}
To assess the inter-modal significance of each patch, we introduce the Inter-Modal Token Selection.
It leverages attention scores from a cross-attention mechanism to identify informative tokens across modalities.
The key insight is to measure the relevance of each patch based on the attention it receives from the remaining modalities.
Specifically, class tokens from all three modalities are concatenated to form the query tokens, while patch tokens are concatenated to form the key tokens.
Then, we can compute the importance scores as follows:
\begin{equation}
Q = \mathcal{T} [f_R^c,f_N^c,f_T^c],
\end{equation}
\begin{equation}
K = \mathcal{C} [f_R^p,f_N^p,f_T^p],
\end{equation}
\begin{equation}
\mathcal{S} = \mathrm{Softmax}\left(\frac{QK^T}{\sqrt{D}}\right),
\end{equation}
where $Q\in \mathbb{R}^{3\times D}$, $K\in \mathbb{R}^{3 L \times D}$ and $\mathcal{S}\in \mathbb{R}^{3\times 3L}$.
Here, $\mathcal{T}$ denotes a stacking operation followed by a linear projection, and $\mathcal{C}$ denotes the concatenation followed by a linear layer.
Then, we separate the scores of each modality from $\mathcal{S}$:
\begin{equation}
\mathcal{D}_m  = \bar{\mathcal{C}}[\mathcal{S}[u \neq m]],
\end{equation}
where $u\in \{N,R,T\}$, $\bar{\mathcal{C}}$ denotes the concatenation.
$\mathcal{D}_m$ aggregates cross-attention scores by excluding self-modality tokens, indicating the relevance degree of each patch receives from other modalities.
Based on $\mathcal{D}_m$, we can select the most relevant patches from the other modalities as:
\begin{equation}
\bar{\Theta}_m = TopK(\mathcal{D}_m,k_2).
\end{equation}
Here, $k_2$ denotes the number of important patches selected based on cross-modal information and $\bar{\Theta}_m$ represents the corresponding index set.
We then aggregate the patch indices selected by other modalities.
Finally, masks for each modality are constructed as follows:
\begin{equation}
M_m^c = \varPsi(\bar{\Theta}_m).
\end{equation}

To select patches important to both their own modality and others, we combine the intra-modal and inter-modal selection masks via a union operation as follows:
\begin{equation}
M_m = M_m^c \cup M_m^s,
\end{equation}
where $\cup$ denotes the union operation.
Finally, we apply the mask \( M^m \) to the patch tokens \( f_m^p \) to select tokens as follows:
\begin{equation}
f_m^{sp}  = M_m \odot f_m^p.
\end{equation}
Here, $\odot$ means the element-wise multiplication.
Through the above steps, we obtain the selected patch tokens \( f_m^{sp} \) for each modality, which effectively mitigates background interference with both intra-modal and inter-modal information.
\begin{figure}[t]
\centering
\includegraphics[width=0.9\columnwidth]{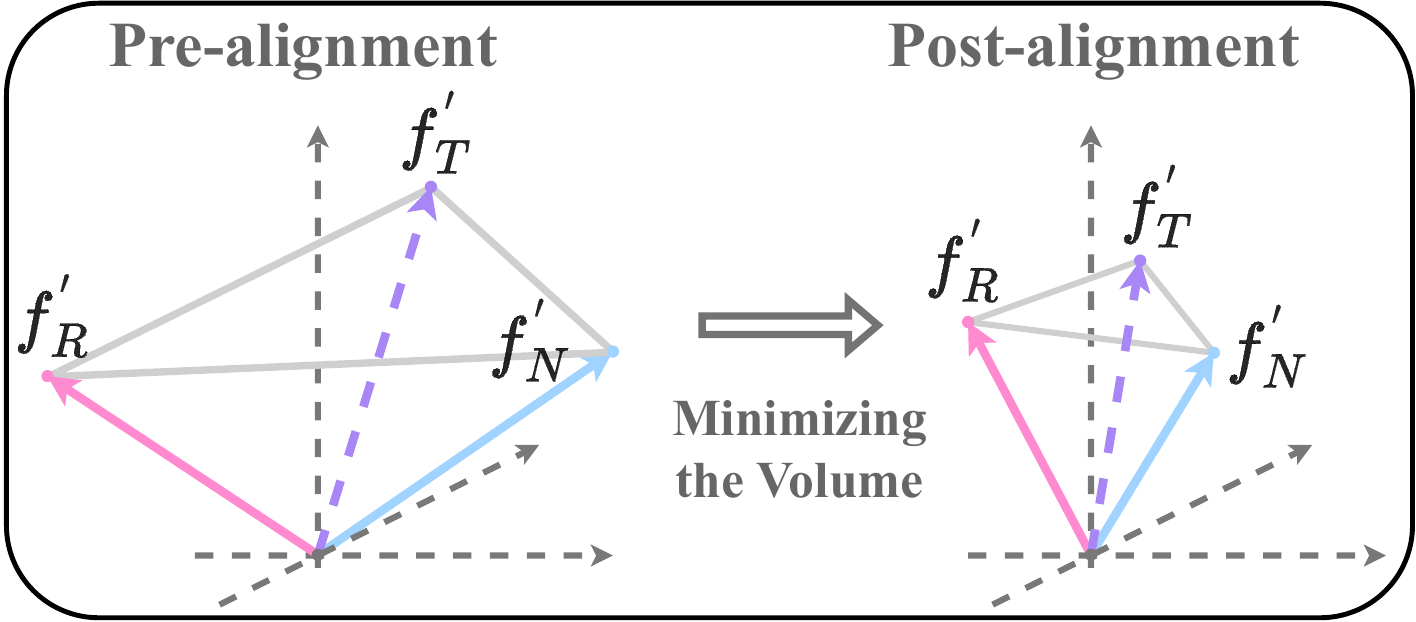}
\caption{Details of Global Alignment Module.}
\label{fig:global}
\end{figure}

\noindent\textbf{Modal Interaction.}
To further reduce the background interference, we propose a modal interaction module that highlights informative features.
It utilizes multi-head cross-attention~\cite{vaswani2017attention} and a feed-forward network to model interactions between the selected tokens and class tokens, thereby extracting more discriminative representations.
Specifically, we concatenate the class tokens \( f_m^c \) and selected patch tokens \( f_m^{sp} \) to form a query \( \bar{Q} \) and a key \( \bar{K} \):
\begin{equation}
\bar{Q} = \mathcal{T}[f_R^c,f_N^c,f_T^c], \bar{K} = \mathcal{C}[f_{R}^{sp},f_{N}^{sp},f_{T}^{sp}],
\end{equation}
where \( \bar{Q} \in \mathbb{R}^{3 \times D} \) and \( \bar{K} \in \mathbb{R}^{3L \times D} \).
Then, we apply multi-head cross-attention to enhance the interaction between the class tokens and selected patch tokens as follows:
\begin{equation}
\bar{Q}' = LN(\bar{Q} + MHCA(\bar{Q},\bar{K},\bar{K})),
\end{equation}
\begin{equation}
f_{rnt} = LN(\bar{Q}' + \phi(\bar{Q}')).
\end{equation}
where $MHCA$ represents the multi-head cross-attention.
$LN$ represents the layer normalization~\cite{ba2016layernorm}.
$\phi(\cdot)$ means the feed-forward network.
Ultimately, we obtain the modality interaction feature \( f_{rnt} \in \mathbb{R}^{3D} \), which aggregates the discriminative information from the selected patch tokens and class tokens across all modalities.
\subsection{Global Alignment Module}
Multi-modal alignment encourages consistent semantic representations across modalities, reducing cross-modal conflicts.
However, traditional alignment methods~\cite{ruan2023clip4vla,girdhar2023imagebind,chen2023vast} are difficult to extend to multiple modalities.
Otherwise it leads to extremely high complexity.
Meanwhile, simultaneous alignment across multiple modalities remains unexplored in the context of multi-modal object ReID.
Motivated by these observations, we introduce Global Alignment Module (GAM) based on multi-modal representation learning in the gramian space~\cite{cicchetti2025gramian}.
It addresses these limitations by ensuring that all modalities are aligned with each another, rather than merely aligning each modality to a designated anchor.
Specifically, we preprocess patch tokens as follows:
\begin{equation}
f_m = Mean(f_m^p),
\end{equation}
\begin{equation}
f_m' = \frac{f_m}{||f_m||_2}.
\end{equation}
Here, $f_m$ denotes the average feature vector.
$f_m'$ is the normalized vector.
As shown in Fig.~\ref{fig:global}, we consider the volume of a 3D polyhedron composed of three vectors (i.e.,  $f_R'$, $f_N'$ and $f_T'$) as a measure of the alignment degree of the three modalities.
A larger volume indicates a worse alignment of the three modalities, while a smaller volume indicates a better alignment.
We interpret the alignment quality of the modalities through this volume metric.
Therefore, we can calculate the volume to align the three vectors.
More specifically, $f_R'$, $f_N'$ and $f_T'$ can be arranged into columns of a matrix \textbf{A} = ($f_R'$, $f_N'$, $f_T'$).
The Gram matrix $\mathbf{G}(f_R', f_N', f_T')$ is defined as:
\begin{align}
\mathbf{G}(f_R', f_N', f_T') &= \mathbf{A}^\top \mathbf{A}, \label{eq:G1} \\
&=
\begin{bmatrix}
\langle \mathbf{f_R'}, \mathbf{f_R'} \rangle & \langle \mathbf{f_R'}, \mathbf{f_N'} \rangle & \langle \mathbf{f_R'}, \mathbf{f_T'} \rangle \\
\langle \mathbf{f_N'}, \mathbf{f_R'} \rangle & \langle \mathbf{f_N'}, \mathbf{f_N'} \rangle & \langle \mathbf{f_N'}, \mathbf{f_T'} \rangle \\
\langle \mathbf{f_T'}, \mathbf{f_R'} \rangle & \langle \mathbf{f_T'}, \mathbf{f_N'} \rangle & \langle \mathbf{f_T'}, \mathbf{f_T'} \rangle
\end{bmatrix}, \label{eq:G2}
\end{align}
where $\mathbf{G}(f_R', f_N', f_T')\in \mathbb{R}^{3 \times 3}$ is the geometric relationship among the three vectors.
The volume of the 3D polyhedron spanned by these vectors is:
\begin{equation}
\operatorname{Vol}(f_R', f_N', f_T') = \sqrt{\det \mathbf{G}(f_R', f_N', f_T')}.
\end{equation}
Here, $\operatorname{Vol}$ is the volume and $\det\mathbf{G}$ represents the determinant of matrix $\mathbf{G}$.
Minimizing this volume enables simultaneous alignment of the tri-modal features, thereby achieving global alignment.
In contrast to traditional alignment methods, GAM exhibits superior efficiency in the simultaneous alignment of an arbitrary number of modalities.
\begin{figure}[t]
\centering
\includegraphics[width=1.0\columnwidth]{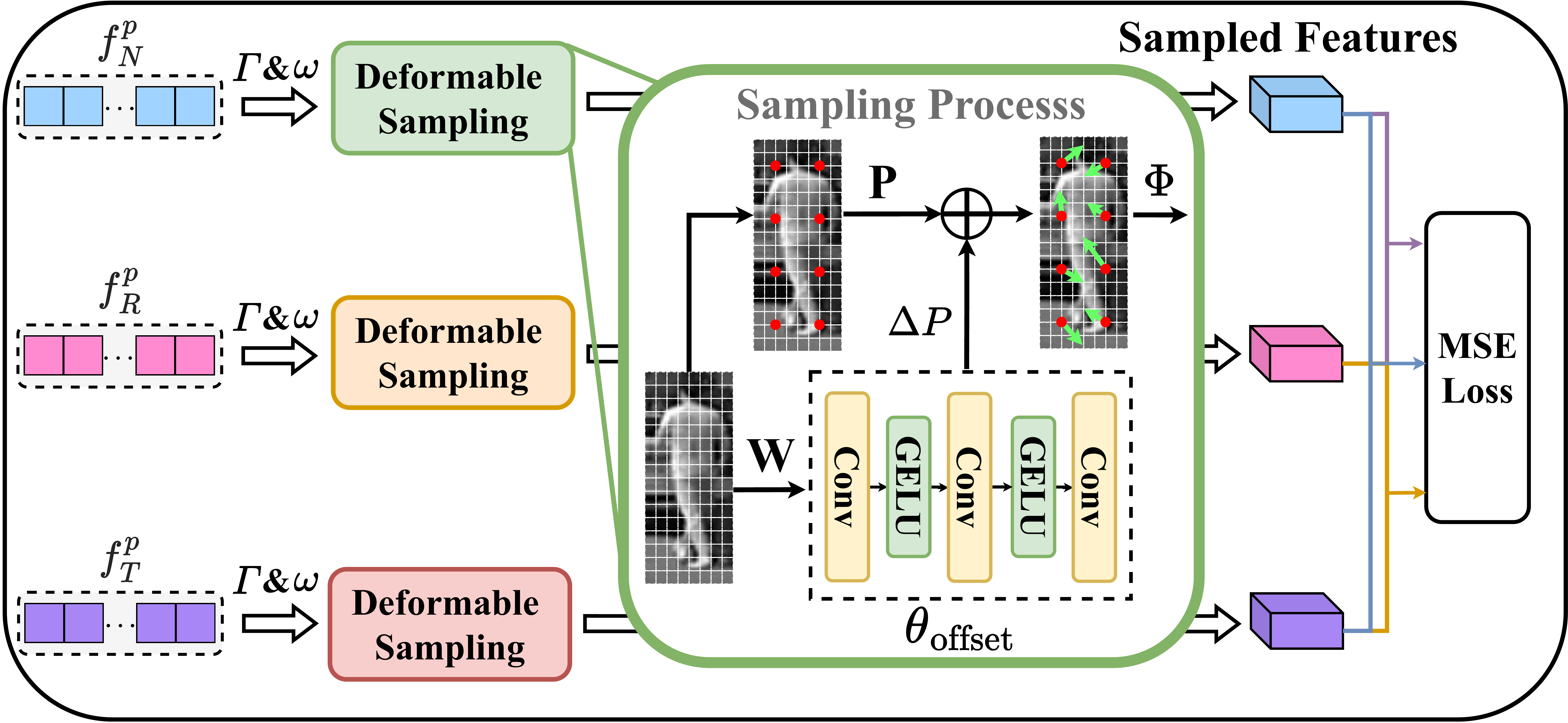}
\caption{Mechanism of Local Alignment Module.}
\label{fig:local}
\end{figure}
\subsection{Local Alignment Module}
The GAM achieves global alignment among the three modalities, while neglecting the issue of pixel misalignment in multi-modal imaging.
To address this issue, we propose the Local Alignment Module (LAM).
Unlike traditional local alignment methods~\cite{wang2024attribute,li2025alignmamba,wang2022multi}, LAM emphasizes adaptive offset sampling and focuses on key details.
As shown in Fig.~\ref{fig:local}, we utilize the advantages of deformable attention~\cite{xia2022vision} by learning the offset to automate the correction of pixel offset errors.
Specifically, we reshape the patch tokens into the spatial manner and generate uniform grid points as follows:
\begin{equation}
P = \omega (\varGamma (f_m^{p})),
\end{equation}
where $\varGamma$ represents the reshape operation.
$\omega$ denotes generating a uniform grid of points and {\em $P \in \mathbb{R}^{H_g\times W_g \times 2}$} are reference points.
The grid size is down-sampled by a factor {\em $r, H_g = \frac{H}{r},W_g = \frac{W}{r}$.}
The values of the reference points are linearly spaced at 2D coordinates $\{(0,0),\ldots,(H_g-1,W_g-1)\}$ and then normalized to the range $[-1, +1]$ based on the grid shape $H_g \times W_g$, where $(-1, -1)$
represents the top-left corner and $(+1, +1)$ represents the bottom-right corner.
To obtain the offset, we perform the following operation on $f_m^{p}$:
\begin{equation}
\Delta P = \theta_{\text{offset}}(f_m^{p}W),
\end{equation}
where $W$ represents a linear projection layer, $\theta_{\text{offset}}(\cdot)$ is composed of multiple convolutional layers.
Here, \( \Delta P \) denotes the sampling offset, which adjusts the reference positions \( P \) in the feature map.
The sampled features \( \bar{f_m^{p}} \) are then obtained using a bilinear interpolation function \( \Phi(\cdot) \), which extracts features from positions \( P + \Delta P \) as follows:
\begin{equation}
\bar{f_m^{p}} = \Phi(f_m^{p}; P + \Delta P).
\end{equation}

After obtaining the sampled features of three modalities, we align these features using the MSE loss to facilitate cross-modal feature alignment.
Through the integration of LAM, our model exhibits the capability of mitigating pixel-level misalignment across different modalities, thereby enhancing the semantic consistency of multi-modal features.
\subsection{Objective Functions}
As shown in Fig.~\ref{fig:Signal}, we optimize the framework using multiple losses.
Features after SIM are supervised by the label smoothing cross-entropy loss~\cite{szegedy2016rethinking}
and triplet loss~\cite{hermans2017defense}:
\begin{equation}
\mathcal{L}_g = \mathcal{L}_{CE} + \mathcal{L}_{TRI},
\end{equation}
where $\mathcal{L}_{CE}$ is the label smoothing cross-entropy loss.
$\mathcal{L}_{TRI}$ is the triplet loss.
Features after GAM are supervised by the gram multi-modal contrastive loss~\cite{cicchetti2025gramian}:
\begin{equation}
\scalebox{0.8}{$\displaystyle
\begin{aligned}
\mathcal{L}_{D2A} &= -\frac{1}{B} \sum_{i=1}^{B} \log \frac{\exp\bigl(-\text{Vol}(\mathbf{a}_i, \mathbf{m}_{2i}, \mathbf{m}_{3i}) / \tau\bigr)}{\sum_{j=1}^{K} \exp\bigl(-\text{Vol}(\mathbf{a}_j, \mathbf{m}_{2i}, \mathbf{m}_{3i}) / \tau\bigr)},
\end{aligned}$}
\end{equation}
\begin{equation}
\scalebox{0.8}{$\displaystyle
\begin{aligned}
\mathcal{L}_{A2D} &= -\frac{1}{B} \sum_{i=1}^{B} \log \frac{\exp\bigl(-\text{Vol}(\mathbf{a}_i, \mathbf{m}_{2i}, \mathbf{m}_{3i}) / \tau\bigr)}{\sum_{j=1}^{K} \exp\bigl(-\text{Vol}(\mathbf{a}_i, \mathbf{m}_{2j}, \mathbf{m}_{3j}) / \tau\bigr)}.
\end{aligned}$}
\end{equation}
Here, $B$ is the batch size, $K$ is the number of modalities, $a_x$ refers to the embeddings of the anchor modality of the $x$-th sample in the batch,
while $m_{xy}$ refers to the embedding of the $x$-th modality of the $j$-th sample in the batch,
$\tau$ is a learnable scaling parameter.
As for LAM, the output features are supervised by the MSE loss as follows:
\begin{equation}
\mathcal{L}_{MSE} = \frac{1}{n} \sum_{i=1}^{n} (x_i - \hat{x}_i)^2.
\end{equation}
Here, $x_i$ is the true value, $\hat{x}_i$ is the predict value.
Finally, the overall loss $\mathcal{L}$ for our framework can be given by:
\begin{equation}
\mathcal{L} = \mathcal{L}_g + \alpha (\mathcal{L}_{D2A} + \mathcal{L}_{A2D}) + \beta\mathcal{L}_{MSE}.
\end{equation}
\section{Experiments}
\subsection{Datasets and Evaluation Metrics}
\textbf{Datasets.} We evaluate the proposed method on three multi-modal object ReID benchmarks.
RGBNT201~\cite{zheng2021robust} is a multi-modal person ReID dataset.
It contains 4,787 RGB, NIR and TIR image triples, captured from 201 distinct identities.
RGBNT100~\cite{li2020multi} is a large scale multi-modal vehicle ReID dataset.
It comprises 17,250 image triples and encompasses a broad spectrum of visual scenarios.
MSVR310~\cite{zheng2022multi} serves as a small scale multi-modal vehicle ReID dataset.
It has 2,087 high quality image triples taken in various environments.

\textbf{Evaluation Metrics.} To evaluate the performance, we utilize the mean Average Precision (mAP)
and Cumulative Matching Characteristics (CMC) at Rank-K ( K$ = 1, 5, 10$).
\subsection{Implementation Details}
The proposed model is implemented in PyTorch and trained using two NVIDIA GeForce RTX 3090 GPUs.
We use the pre-trained CLIP~\cite{radford2021learning} as the visual encoder.
Images in triples are resized to 256×128 for RGBNT201, 128×256 for RGBNT100 and MSVR310.
For data augmentation, we apply random horizontal flipping, cropping and erasing~\cite{zhong2020random}.
For RGBNT201 and MSVR310, the mini-batch size is set to 64, sampling 8 and 4 images per identity respectively.
For RGBNT100, the mini-batch size is 128 with 16 images per identity.
We use the Adam optimizer~\cite{Kinga2015Adam} to fine-tune the proposed modules with a learning rate of $3.5\text{e}^{-4}$ and the visual encoder with a relatively low learning rate of $5\text{e}^{-6}$.
We train the model for 50 epochs.
\begin{table}[t]
\centering
\resizebox{0.9\linewidth}{!}{
\begin{tabular}{ccccc}
\toprule
\textbf{Methods} & \textbf{mAP} & \textbf{R-1} & \textbf{R-5} & \textbf{R-10} \\
\midrule
HAMNet~\cite{li2020multi}    & 27.7 & 26.3 & 41.5 & 51.7 \\
PFNet~\cite{zheng2021robust}  & 38.5 & 38.9 & 52.0 & 58.4 \\
DENet~\cite{zheng2023dynamic}  & 42.4 & 42.2 & 55.3 & 64.5 \\
IEEE~\cite{wang2022interact}   & 47.5 & 44.4 & 57.1 & 63.6 \\
LRMM~\cite{wu2025lrmm}      & 52.3 & 53.4 & 64.6 & 73.2 \\
UniCat$^*$~\cite{crawford2023unicat} & 57.0 & 55.7 & -    & -    \\
HTT$^*$~\cite{wang2024heterogeneous}   & 71.1 & 73.4 & 83.1 & 87.3 \\
TOP-ReID$^*$~\cite{wang2024topreid} & 72.3 & 76.6 & 84.7 & 89.4 \\
EDITOR$^*$~\cite{zhang2024magic} & 66.5 & 68.3 & 81.1 & 88.2 \\
RSCNet$^*$~\cite{yu2024representation}  & 68.2 & 72.5 & -    & -    \\
DeMo$\dagger$~\cite{wang2025demo}  &79.0  &\underline{82.3} &88.8 &92.0 \\
IDEA$\dagger$~\cite{wang2025idea}  &\underline{80.2} &82.1 &\underline{90.0} &\underline{93.3} \\
PromptMA$\dagger$~\cite{zhang2025promptma}  &78.4 &80.9 &87.0 &88.9 \\
\textbf{Signal$\dagger$ (Ours)}       &\textbf{80.3} &\textbf{85.2} &\textbf{91.4} &\textbf{93.7} \\
\bottomrule
\end{tabular}
}
\caption{Performance comparison on RGBNT201. The best and second results are in bold and underlined, respectively. The symbol $\dagger$ denotes CLIP-based methods, $*$ indicates ViT-based methods and others are CNN-based methods.}
\label{tab:RGBNT201}
\end{table}
\begin{table}[t]
\centering
\resizebox{0.9\linewidth}{!}{
\begin{tabular}{cccccccc}
\toprule
& \multirow{2}{*}{\textbf{Methods}} & \multicolumn{2}{c}{\textbf{RGBNT100}} & \multicolumn{2}{c}{\textbf{MSVR310}} \\
\cmidrule(lr){3-4} \cmidrule(lr){5-6}
& & \textbf{mAP} & \textbf{R-1} & \textbf{mAP} & \textbf{R-1} \\
\midrule
& GAFNet~\cite{guo2022generative}    & 74.4 & 93.4 & -    & -    \\
& GPFNet~\cite{he2023graph}    & 75.0 & 94.5 & -    & -    \\
& PFNet~\cite{zheng2021robust}   & 68.1 & 94.1 & 23.5 & 37.4 \\
& HAMNet~\cite{li2020multi}     & 74.5 & 93.3 & 27.1 & 42.3 \\
& CCNet~\cite{zheng2022multi}   & 77.2 & 96.3 & 36.4 & 55.2 \\
& LRMM~\cite{wu2025lrmm}       & 78.6 & 96.7 & 36.7 & 49.7 \\
& PHT$^*$~\cite{pan2023progressively}      & 79.9 & 92.7 & -    & -    \\
& HTT$^*$~\cite{wang2024heterogeneous}    & 75.7 & 92.6 & -    & -    \\
& TOP-ReID$^*$~\cite{wang2024topreid}& 81.2 & 96.4 & 35.9 & 44.6 \\
& EDITOR$^*$~\cite{zhang2024magic}& 82.1 & 96.4 & 39.0 & 49.3 \\
& RSCNet$^*$~\cite{yu2024representation}   & 82.3 & 96.6 & 39.5 & 49.6 \\
&DeMo$\dagger$~\cite{wang2025demo}  &86.2  &97.6 &49.2 &59.8 \\
&IDEA$\dagger$~\cite{wang2025idea}  &\textbf{87.2} &96.5 &47.0 &62.4 \\
&PromptMA$\dagger$~\cite{zhang2025promptma}  &85.3 &\underline{97.4} &\textbf{55.2} &\underline{64.5} \\
&\textbf{Signal$\dagger$ (Ours)}       &\underline{86.3} &\textbf{97.6} &\underline{53.6} &\textbf{71.9} \\
\bottomrule
\end{tabular}
}
\caption{Performance on RGBNT100 and MSVR310.}
\label{tab:RGBNT100andMSVR310}
\end{table}
\subsection{Comparison with State-of-the-Art Methods}
\textbf{Multi-modal Person ReID.}
In Tab.~\ref{tab:RGBNT201}, we compare our method with other methods on RGBNT201.
Generally, multi-modal methods show a considerable improvement over single-modal methods by incorporating complementary information.
Among these methods, models based on CLIP perform better.
Specifically, our framework improves by 3.1\% in Rank-1 accuracy compared to IDEA.
This highlights the effectiveness of hierarchical alignment of different modalities.
Besides, compared to DeMo, our method improves by 1.3\% in mAP.
These results confirm the effectiveness of our framework for multi-modal person ReID.

\noindent\textbf{Multi-modal Vehicle ReID.}
In Tab.~\ref{tab:RGBNT100andMSVR310}, we compare our method with other methods on the RGBNT100 and MSVR310 datasets.
on RGBNT100, our method improves Rank-1 by 1.1\% compared to IDEA.
In comparison with EDITOR, our method achieves a 4.2\% improvement in mAP and a 1.2\% improvement in Rank-1.
On MSVR310, our method improves mAP by 6.6\% and Rank-1 by 9.5\% compared to IDEA.
These results indicate the effectiveness of our framework for multi-modal vehicle ReID.
\begin{table}[t]
    \centering
    \renewcommand\arraystretch{1.0}
    \setlength\tabcolsep{4.5pt}
    \resizebox{0.40\textwidth}{!}
    {
    \begin{tabular}{ccccccccc}
        \noalign{\hrule height 1pt}
        \multicolumn{1}{c}{\multirow{2}{*}{\textbf{Models}}}                   &\multicolumn{3}{c}{\textbf{Modules}}                                  & \multicolumn{2}{c}{\textbf{Metrics}} & \multicolumn{2}{c}{\textbf{Params}}     \\
        \cmidrule(r){2-4} \cmidrule(r){5-6} \cmidrule(r){7-7} \cmidrule(r){8-8}
    & \textbf{SIM}              & \textbf{GAM}                & \textbf{LAM}                   & \textbf{mAP}    & \textbf{R-1}   & \textbf{M} & $\uparrow$\textbf{\%}                 \\ \hline
    A                  & \ding{53}                  & \ding{53}                  & \ding{53}                    & 70.3  & 71.8 & 86.41  & -                 \\
    B                  & \ding{51}                  & \ding{53}                  & \ding{53}                      & 77.0  & 80.6 & 89.56 & 3.65                 \\
    \multirow{1}{*}{C} & \multirow{1}{*}{\ding{51}} & \multirow{1}{*}{\ding{51}} & \multirow{1}{*}{\ding{53}}    & 79.0 & 82.8 & 89.56    &  3.65            \\
    \multirow{1}{*}{D} & \multirow{1}{*}{\ding{51}} & \multirow{1}{*}{\ding{51}} & \multirow{1}{*}{\ding{51}}    & \textbf{80.3}  & \textbf{85.2} & \textbf{91.17}   & \textbf{5.51}             \\
    \noalign{\hrule height 1pt}
    \end{tabular}
    }
    \caption{Comparison with different modules on RGBNT201.}
    \label{tab:different module}
\end{table}
\begin{figure}[t]
\centering
\includegraphics[width=0.8\columnwidth,height=0.64\columnwidth]{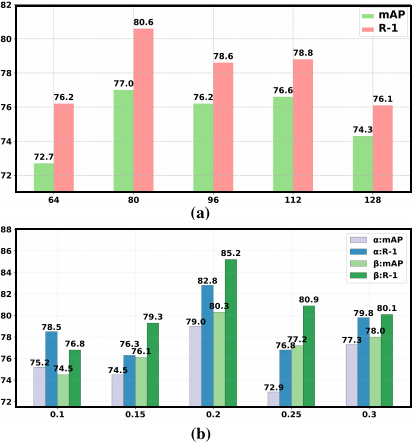}
\caption{
    (a) Performance with different numbers of reserved tokens.
    (b) Performance with different $\alpha$ and $\beta$. }
\label{fig:params}
\end{figure}
\subsection{Ablation Studies}
We evaluate the effectiveness of different modules on the RGBNT201 dataset.
Our baseline employs a concatenated feature of tri-modal class tokens from the visual encoder.

\noindent\textbf{Effects of Key Modules.}
Tab.~\ref{tab:different module} shows the performance comparison of different modules.
Model A is the baseline model, achieving 70.3\% mAP and 71.8\% Rank-1 accuracy.
After adding SIM, the performance of Model B improves to 77.0\% mAP and 80.6\% Rank-1.
This indicates that removing useless image backgrounds is of great significance.
Model C further integrates GAM, increasing mAP to 79.0\% and Rank-1 to 82.8\%,
demonstrating the effectiveness of the modality alignment.
Finally, Model D combines all modules and achieves the best results,
with 80.3\% mAP and 85.2\% Rank-1.
The complexity analysis shows that our proposed modules add only \textbf{4.76MB} learnable parameters in total.
It achieves great performance gains with fewer parameters.

\noindent\textbf{Effects of the Number of Reserved Tokens.}
Fig.~\ref{fig:params} (a) demonstrates how the number of reserved patch tokens (\( k_1 \)) affects the retrieval performance.
The best result is observed when \( k_1 = 80 \).
Thus, we set \( k_1 = 80 \) as default.

\noindent\textbf{Effects of Loss Weights on GAM and LAM.}
Fig.~\ref{fig:params} (b) presents ablation results of loss weights.
When $\alpha$ is 0.2, adding GAM increases the mAP of the model to 79.0\% and Rank-1 to 82.8\%.
When $\beta$ is 0.2, adding LAM increases the mAP of the model to 80.3\% and Rank-1 to 85.2\%.
Thus, we utilize these optimal weights as default settings.

\noindent\textbf{Effect of Mask Intersection and Union.}
Tab.~\ref{tab:Intersection or Union} presents a comparison investigation for the intersection and union of $M_m^c$ and $M_m^s$  within SIM.
The results reveal that the union operation yields higher performances.
The union operation delivers 1.1\% higher in mAP than the intersection operation.

\noindent\textbf{Effect of Offset Sharing and Non-Sharing.}
Tab.~\ref{tab:offset sharing or no-sharing} presents a comparative analysis with the offset sharing and non-sharing among RGB, NIR and TIR modalities in the LAM.
Compared with the sharing offset, the non-sharing offset yields a 5.7\% improvement in mAP and a 6.9\% improvement in Rank-1.
This indicates that the utilization of independent offsets for each modality exhibits greater efficacy.
\begin{figure*}[t]
\centering
\includegraphics[width=0.9\textwidth,height=0.44\columnwidth]{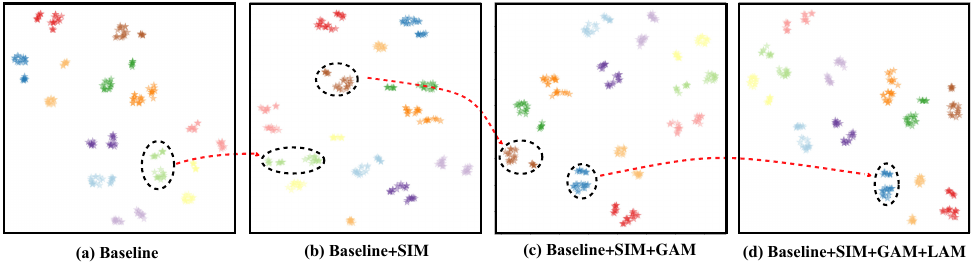}
\caption{Visualization of the feature distributions with t-SNE. Different colors stand for different identities.}
\label{fig:tsne}
\end{figure*}
\begin{table}[t]
\centering
\fontsize{6pt}{4pt}\selectfont
\resizebox{0.9\linewidth}{!}{
\begin{tabular}{ccccc}
\toprule
\textbf{Methods} & \textbf{mAP} & \textbf{R-1} & \textbf{R-5} & \textbf{R-10} \\
\midrule
Intersection    & 79.2 & 82.3 & 89.5 & 93.7 \\
\textbf{Union}           &\textbf{80.3} &\textbf{85.2} &\textbf{91.4} &\textbf{93.7} \\
\bottomrule
\end{tabular}
}
\caption{Comparison of the intersection and union in SIM.}
\label{tab:Intersection or Union}
\end{table}
\begin{table}[t]
\centering
\fontsize{6pt}{4pt}\selectfont
\resizebox{0.9\linewidth}{!}{
\begin{tabular}{ccccc}
\toprule
\textbf{Methods} & \textbf{mAP} & \textbf{R-1} & \textbf{R-5} & \textbf{R-10} \\
\midrule
Sharing             & 74.6 & 78.3 & 86.6 & 92.0 \\
\textbf{Non-Sharing}          &\textbf{80.3} &\textbf{85.2} &\textbf{91.4} &\textbf{93.7} \\
\bottomrule
\end{tabular}
}
\caption{Comparison of whether offset is shared in LAM.}
\label{tab:offset sharing or no-sharing}
\end{table}
\subsection{Visualization Analysis}
\noindent\textbf{Multi-modal Feature Distributions with t-SNE.}
Fig.~\ref{fig:tsne} shows the feature distribution with different modules.
Comparing Fig.~\ref{fig:tsne} (a) and (b), as instances with the same ID become more compact, removing redundant patches improves the feature discrimination ability.
In Fig.~\ref{fig:tsne} (c), by using GAM, the feature distribution is more compact than the one in Fig.~\ref{fig:tsne} (b).
In Fig.~\ref{fig:tsne} (d), the feature distribution is more compact than the one in Fig.~\ref{fig:tsne} (c), indicating that LAM enhances feature discrimination.
These results demonstrate the effectiveness of our proposed modules.
\begin{figure}[ht]
\centering
\includegraphics[width=1.0\columnwidth]{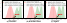}
\caption{Visualization of the cosine similarity distribution.}
\label{fig:cosine}
\end{figure}
\begin{figure}[ht]
\centering
\includegraphics[width=0.8\columnwidth,height=0.34\columnwidth]{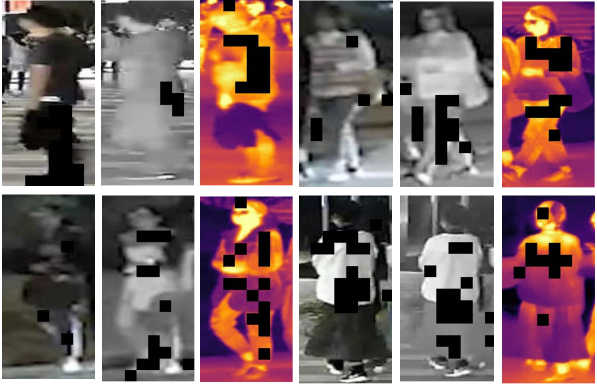}
\caption{Visualization of token selection. Black blocks denote the removed image content.}
\label{fig:tokenvisual}
\end{figure}
\begin{figure}[ht]
\centering
\includegraphics[width=0.8\columnwidth,height=0.34\columnwidth]{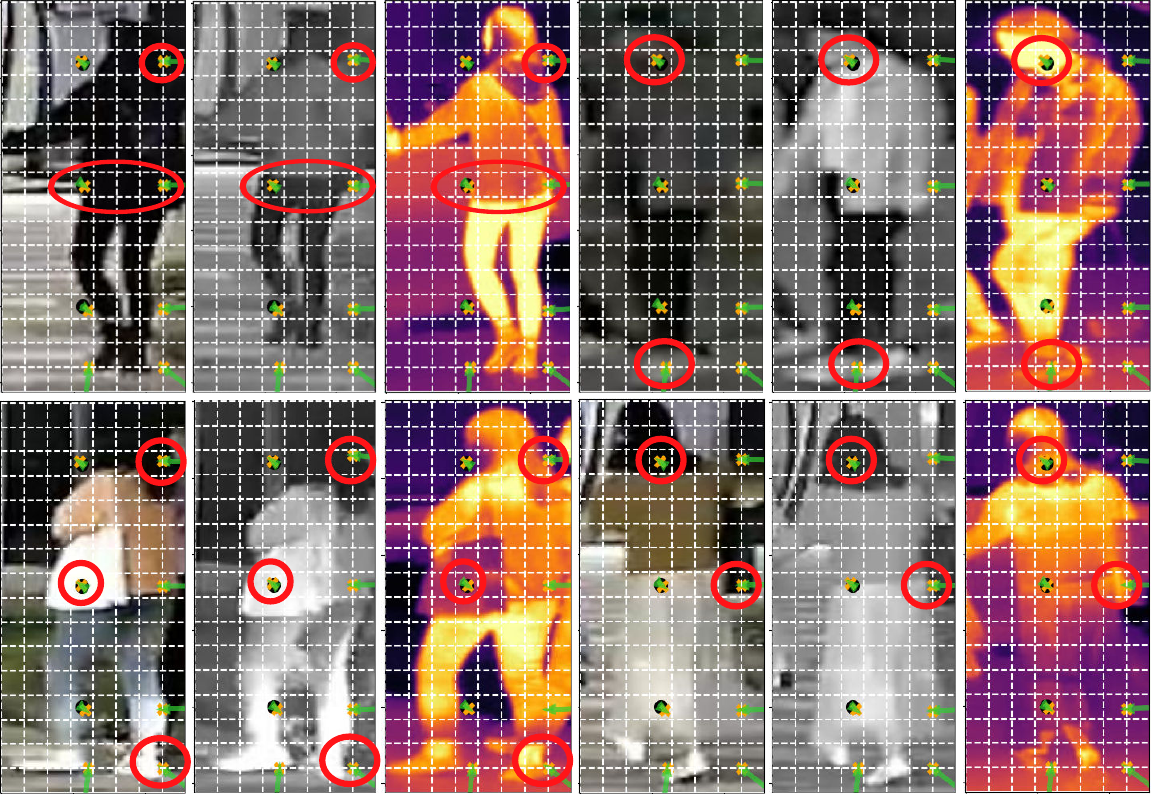}
\caption{The visualization of generated offsets.}
\label{fig:offsets}
\end{figure}

\noindent\textbf{Cosine Similarity Distributions.}
Fig.~\ref{fig:cosine} shows the distributions of cosine similarities among test features.
As observed, the intersected area of the two feature distributions is decreasing.
It indicates our framework further amplifies the discrepancy between positive and negative samples.

\noindent\textbf{Token Selection in SIM.}
As shown in Fig.~\ref{fig:tokenvisual}, some patches are removed from each modality.
This indicates that intra-modal and inter-modal token selection achieve the expected goal of retaining important patch tokens.

\noindent\textbf{Visualization of Generated Offsets.}
Fig.~\ref{fig:offsets} demonstrates the alignment of local details across image triples for each object.
The first one in the second row shows the model's effectiveness in aligning fine-grained details such as hair, shoulder bags and heels across modalities.
It solves the problem of pixel-level shift.
This validates the model's capability to learn local detail alignment.
\section{Conclusion}
In this paper, we propose a novel framework named \textbf{Signal} for multi-modal object ReID.
Our method first uses a Selective Interaction Module (SIM) to select important patch tokens from multi-modal images.
Then, we introduce the Global Alignment Module (GAM) to achieve feature alignment across multiple modalities.
Finally, the Local Alignment Module (LAM) aligns important details within each modality in a shift-aware manner.
As a result, our framework can extract more effective features for multi-modal object ReID.
Extensive experiments on three benchmark datasets validate the effectiveness of our proposed method.
\section{Acknowledgments}
This work was supported in part by the National Natural Science Foundation of China (No.62576069, 62506272), Natural Science Foundation of Liaoning Province (No.2025-MS-025) and Dalian Science and Technology Innovation Fund (No.2023JJ11CG001).
\bibliography{aaai2026}

\end{document}